\documentclass[10pt, a4paper]{article} 
\usepackage{lrec2026}
\usepackage{xcolor}
\usepackage{booktabs}
\usepackage{comment}
\usepackage{amsfonts}
\usepackage{amsmath}
\usepackage{multirow}
\usepackage{placeins}
\usepackage{times}
\usepackage{latexsym}

\newcommand{\lb}[1]{\textcolor{blue}{#1}}
\newcommand{\ssh}[1]{\textcolor{red}{#1}}

\usepackage{soul}
\usepackage{enumitem}

\title{Can LLM Reasoning Be Trusted? A Comparative Study: Using Human Benchmarking on Statistical Tasks}

\name{Crish Nagarkar\textsuperscript{\rm 1}, Leonid Bogachev\textsuperscript{\rm 1}, Serge Sharoff\:\!\textsuperscript{\rm 2}}

\address{\\[-.6pc]
\textsuperscript{\rm 1}\:\!Department of Statistics, School of Mathematics\\
\textsuperscript{\rm 2}\:\!School of Languages, Cultures \& Societies\\
University of Leeds, Leeds LS2 9JT, UK\\
\{C.A.Nagarkar, L.V.Bogachev, S.Sharoff\}@leeds.ac.uk\\[-.6pc]}

\abstract{
This paper investigates the ability of large language models (LLMs) to solve statistical tasks, as well as their capacity to assess the quality of 
reasoning. While state-of-the-art LLMs have demonstrated remarkable performance in a range of NLP tasks, their 
competence in addressing even moderately complex statistical challenges 
is not well understood.
We have fine-tuned selected open-source LLMs 
on a specially developed dataset to enhance their statistical reasoning capabilities, and compared their performance with the human scores used as a benchmark. 
Our results show that the fine-tuned models achieve better performance on advanced statistical tasks on the level comparable to a statistics student. Fine-tuning demonstrates architecture-dependent improvements, with some models showing significant performance gains, indicating clear potential for deployment in educational technology and statistical analysis assistance systems.
We also show that LLMs themselves can be far better judges of the answers quality (including explanation and reasoning assessment) in comparison to traditional metrics, such as BLEU or BertScore.
This self-evaluation capability enables scalable automated assessment for statistical education platforms and quality assurance in automated analysis tools. Potential applications also include 
validation tools for research methodology in academic and industry settings, and quality control mechanisms for data analysis workflows.
\\[.2pc]
\Keywords 
evaluation methodologies, LLM-as-a-Judge, automated evaluation metrics, statistical problem solving, model reliability
}

\begin{document}
\maketitleabstract

\setlist{itemsep=2pt,parsep=1pt}

\section{Introduction}

Large Language Models (LLMs), both proprietary and open-source ones, such as LLaMA, DeepSeek and Mistral, have demonstrated impressive capabilities in general NLP tasks, including text generation, summarization, and logical inference \citep{touvron2023llama, deepseek2025r1,mistral2023}. However, their ability to perform reasoning for specific tasks remains an open question, since their representations are fine-tuned to predict the next word, with higher-level reasoning existing only as an emergent property, which often leads to limited capabilities \citep{wang2024causalbench}.

This also applies to the specific task tested in this study, namely statistical reasoning and problem solving, such as computing probabilities, interpreting regression results, or applying hypothesis testing \citep{frieder2023mathematical, hendrycks2021}.  Statistical problem-solving requires precise numerical computation, deep conceptual understanding, and logical derivation\,---\,all of which differ markedly from the pattern-matching strengths of pretrained LLMs \citep{bommasani2021opportunities}. Prior work has shown that even state-\allowbreak{}of-\allowbreak{}the-\allowbreak{}art models struggle with the following hurdles:

\begin{itemize}
[noitemsep]
\item \textbf{Computational inconsistencies}: e.g., miscalculating $p$-values or probabilities \citep{razeghi2022impact}.
\item \textbf{Overfitting to seen templates}: poor generalization to reworded or novel statistical questions \citep{bubeck2023sparks}.
\item \textbf{Conceptual confusion}: failing to distinguish similar concepts like Type I error vs.\ significance level.
\end{itemize}

Our research questions are:
\begin{description}
[noitemsep]
    \item[\textbf{RQ1:}] Can parameter-efficient fine-tuning (e.g., LoRA) improve the statistical reasoning capabilities of base LLMs?
\item [\textbf{RQ2:}] Which automatic metrics most reliably align with human judgments of correctness?
\item[\textbf{RQ3:}] What are the key failure modes and generalization limitations of fine-tuned models in statistical reasoning, and where do automated metrics systematically diverge from human evaluation?
\end{description}

To explore these, we fine-tune several 7B models on a curated dataset of statistical 
questions using LoRA with 8-bit quantization. We evaluate models on a held-out sample of 50 diverse 
questions across three reasoning dimensions: correctness, step-by-step logic, and explanation quality.

In addition to human scoring, we compute several automated scoring metrics for each model output. We then assess how well these metrics correlate with human ratings. This allows us to evaluate not just model performance, but also the validity of automatic metrics for evaluating statistical reasoning.

Our contributions from this study are:\footnote{The dataset and the respective evaluation tools are available from: \url{https://github.com/crishnagarkarleeds/statistics-llm}.}

\begin{enumerate}
    \item [\bf (i)] \textbf{A comprehensive benchmark dataset} of 2{,}\:\!000 statistical questions
    spanning hypothesis testing, regression, probability theory, and descriptive statistics\,---\,publicly released to advance reproducible research in mathematical reasoning evaluation.

\item [\bf (ii)] \textbf{Fine-tuned statistical reasoning models} based on three 7B-parameter architectures (LLaMA-3, Mistral-Nemo, DeepSeek), demonstrating significant improvements in structured quantitative problem-solving over base models.

\item [\bf (iii)] \textbf{Multi-dimensional evaluation framework} comparing automated metrics (BLEU, BERTScore, SBERT) against LLM-as-judge approaches across correctness, explanation quality, and reasoning dimensions, revealing systematic alignment patterns with human evaluation.

\item [\bf (iv)] \textbf{Empirical insights into model limitations and metric reliability}, including identification of failure modes in complex statistical reasoning and divergence patterns between automated metrics and human judgment in mathematical domains.
\end{enumerate}

\section{Related Work}

\subsection{Foundations of LLMs}
Modern LLMs leverage transformer architectures \citep{vaswani2017attention} to process sequences in parallel and capture contextual relationships across tokens effectively. The most recent generation of LLMs are decoder-only models, optimized for generative tasks through autoregressive decoding mechanisms. They have been pretrained on large corpora, measuring in hundreds of billions of words \citep{brown2020fewshot,touvron2023llama}.

To address these limitations, recent studies have explored fine-tuning approaches to adapt LLMs for specialized domains. Techniques such as parameter-efficient fine-tuning (e.g., LoRA) and memory optimization methods (e.g., 8-bit quantization) have shown promise in enhancing model performance without requiring extensive computational resources \citep{hu2021lora, dettmers2022}. However, the impact of these methods on domain-specific tasks remains underexplored, particularly in mathematical reasoning.

This work builds on prior efforts by systematically evaluating the benefits of fine-tuning LLaMA on a curated dataset for statistical reasoning. By contrasting baseline LLMs with fine-tuned versions, we highlight the potential of these techniques to improve performance across varying levels of task complexity.

\subsection{Automatic Evaluation Methods}

Recent advances in automatic evaluation have demonstrated that sophisticated metrics can provide reliable assessment of complex NLP tasks \citep{fomicheva-specia-2019-taking}. Particularly promising is the emergence of LLM-as-a-judge methodology, which has achieved state-of-the-art performance in translation quality assessment, demonstrating correlation with human judgments comparable to the best traditional metrics \citep{bavaresco25judges}. Similarly, LLM-based evaluators have shown strong alignment with expert human ratings in educational contexts, such as non-native essay assessment \citep{yancey-etal-2023-rating}.

These developments suggest that LLM-based evaluation approaches may capture nuanced aspects of reasoning quality that traditional lexical similarity metrics miss, making them particularly suitable for complex domain-specific tasks like statistical reasoning. Building on this foundation, our study implements LLM-as-a-judge methodology specifically adapted for mathematical problem-solving evaluation.

\subsection{Application to Statistical Reasoning}

As emphasized by \citet{hendrycks2021}, while LLMs excel at simpler tasks like basic arithmetic and algebra, they struggle with advanced statistical tasks. For instance, hypothesis testing requires computing $p$-values, which involves probabilistic reasoning and context-specific interpretations. In practical evaluations, LLMs frequently generate syntactically correct outputs but fail to exhibit the statistical rigor necessary for accurate conclusions. These errors often arise from gaps in numerical computation, incomplete reasoning, or inability to handle complex statistical frameworks holistically.

Existing datasets such as MATH \citep{hendrycks2021} and UCI ML Repository \citep{dua2019} provide valuable benchmarks for elementary or intermediate tasks like probability and basic regression. However, these datasets are not designed to evaluate the sophisticated reasoning needed for more advanced statistical tasks. They also lack diversity in question types, such as those requiring both computational precision and conceptual interpretation, which are critical for real-world applications.

To address these limitations, the present study introduces a dataset encompassing hypothesis testing, ANOVA, and chi-squared tests. This dataset prioritizes tasks that combine logical reasoning with detailed numerical computations, offering a more realistic assessment of LLM performance in statistical problem-solving. By focusing on both computational accuracy and interpretive reasoning, it provides a robust framework for identifying LLMs' strengths and weaknesses, paving the way for advancements in fine-tuning and architecture design for domain-specific tasks.
\section{Methodology}

This section introduces a structured evaluation framework tailored to statistical problem-solving using LLMs.
Building on machine learning lifecycle principles\,---\,Data Understanding, Data Preparation, Model Training, and Model Evaluation \citep{goodfellow2016deep}\,---\,we apply this structure to benchmark LLMs such as LLaMA-3, Mistral-7B, and DeepSeek-R1-Qwen 7B on statistical reasoning tasks.

Our framework departs from traditional NLP benchmarks by emphasizing statistical logic, numerical precision, and interpretability. The pipeline includes custom dataset creation, LoRA-based fine-tuning with quantization, and detailed performance analysis based on both human and automatic evaluations. This methodology enables a rigorous comparison of LLMs’ ability to perform regression analysis, hypothesis testing, probability computation, and other statistical tasks in a structured, explainable manner.

\subsection{Data Curation}


To evaluate LLMs on statistical reasoning, we constructed a custom dataset addressing the limitations of existing datasets such as STATLOG \citep{michie1994machine}, the UCI Machine Learning Repository \citep{dua2019}, and MATH \citep{hendrycks2021}, which focus on traditional machine learning tasks or narrow mathematical problem-solving domains. Our dataset was designed to assess both numerical computation and interpretive reasoning, with problems spanning core statistical topics, including probability, hypothesis testing, regression analysis, ANOVA, and time series analysis.
\subsubsection{Dataset Construction}
The dataset was developed using a multi-step process to ensure a balanced distribution of problem types, incorporating real-world statistical applications such as interpreting $p$-values, regression coefficients, and confidence intervals. The questions were categorized into three groups based on their origin:

\begin{description}
    \item[Open-access educational resources] (63\%). Questions in this group were adapted from widely used open-access academic sources, including  \citet{OpenIntro}, \citet{OpenStax}, \citet{OnlineStatBook}, and \citet{Kozak}. These questions were rewritten to maintain academic rigor while avoiding direct replication and ensuring compliance with open licensing requirements.
    \item[Academic textbook adaptations] (34\%). Questions requiring deeper statistical reasoning or advanced concepts were adapted from established academic textbooks, including \citet{Ross}, \citet{DeGroot}, \citet{wasserman2004all}, \citet{Blitzstein}, \citet{Devore}, and \citet{Gelman}. Specialized topics such as canonical-correlation analysis, cross-correlation at lag~1, multiple correlation coefficients, or 
    multivariate statistical inference, which are rarely found in standard coursework, were specifically included to test high-level reasoning skills beyond traditional undergraduate curriculum. All adapted questions were subjected to human verification and revision to ensure mathematical correctness and academic validity.
    \item[Custom-developed questions] (3\%). Custom questions were developed for specialized statistical scenarios not covered in standard educational materials, focusing on advanced applications such as multi-door Monty Hall-\allowbreak style problems, confidence interval interpretations, and complex ANOVA designs for real-world scenarios. Custom questions ensure more comprehensive coverage of advanced statistical concepts while enhancing the originality of our dataset.
\end{description}

This structured approach guarantees that the dataset reflects both foundational and advanced statistical reasoning tasks, ensuring LLMs are evaluated across diverse statistical challenges while maintaining strict adherence to academic integrity and licensing compliance. Table~\ref{tab:dataset_distribution} presents the distribution of question types across the topics and complexity levels, following established difficulty taxonomies in statistical education \citep{wasserman2004all, moore2003introduction}.


\begin{table*}[t]
\centering \small
\begin{tabular}{l|ccc|c}
\toprule
\textbf{Topic}             & \textbf{Basic} & \textbf{Intermediate} & \textbf{Advanced} & \textbf{Avg.\ Words} \\ \midrule
Probability                & 200            & 150                   & 50                & 30.4                        \\
Hypothesis Testing         & 150            & 100                   & 100               & 84.8                        \\
Regression Analysis        & 100            & 150                   & 200               & 47.0                         \\
Time Series Analysis       & 50             & 75                    & 100               & 26.4                        \\
ANOVA \& Bayesian Stats    & 25             & 75                    & 100               & 67.1                        \\
\bottomrule
\end{tabular}
\caption{Question distribution by topic, difficulty, and length (in words).}
\label{tab:dataset_distribution}
\end{table*}

To assess the statistical reasoning capabilities of LLMs, the dataset incorporated two primary question formats:

\begin{description}
    \item[Numerical answer questions:] tasks requiring direct computational outputs (e.g., regression slopes, probabilities, hypothesis test results).
    \item[Free-text questions:] tasks requiring structured reasoning, stepwise calculations, and statistical interpretation.
\end{description}

\begin{figure*}[t]
\centering
\includegraphics[width=0.8\linewidth]{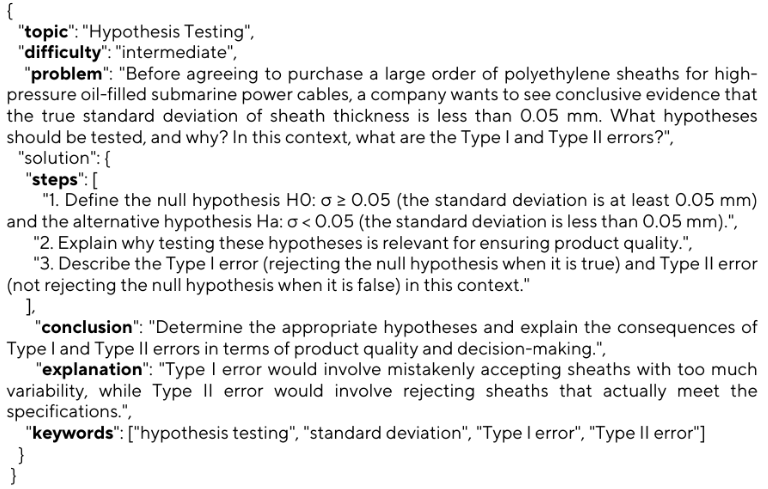}
\caption{Example question with reference solution.}
\label{fig:problem_format}
\end{figure*}

Figure~\ref{fig:problem_format} presents an example question showcasing the structured question format used in the dataset. Using these two formats of questions ensures that LLMs are tested on computational accuracy as well as on their ability to interpret statistical concepts.


\subsection{Evaluation and Marking Scheme}
The evaluation framework emphasized both numerical correctness and conceptual depth, leveraging widely recognized assessment methodologies in AI-driven statistical reasoning \citep{hendrycks2021, cobbe2021gsm8k}.


Each response was graded following a weighted rubric:

\begin{description}
    \item [Correctness of final answer (C)] (40\%): consistency with benchmark results.
    \item [Step-by-step explanation (E)] (35\%): logical structuring and derivation.
    \item [Interpretation and reasoning (R)] (25\%): conceptual clarity and statistical justification.
\end{description}

All models were evaluated on a held-out set of 50 statistical questions by multiple human judges to enable a direct and robust comparison across the different language model architectures. 

This structured evaluation ensures that LLMs are assessed not only for accuracy but also for their ability to demonstrate statistical reasoning, a key challenge in domain-specific AI adaptation \citep{bubeck2023sparks, bommasani2021opportunities}. 

LLM performance was also assessed through several automated metrics: 

\begin{description}
    \item [Perplexity:] evaluates model confidence and coherence in probabilistic text generation\citep{raffel2020exploring}.
    \item [BLEU score:] measures textual similarity between generated answers and reference solutions \citep{papineni2002bleu}.
    \item [SBERT:] measures the distance between generated answers and reference solutions on the basis of SBERT embeddings \citep{reimers2019sentence}.
    \item [BERT score:] a trainable metric for predicting the similarity of generated answers and reference solutions \citep{Zhang20BERTScore}.
    \item [LLM as a Judge:] a prompt presents for each question the reference solution and the output of an LLM with the task of grading it with the same rubric as used by human raters  \citep{kocmi-federmann-2023-large}.
\end{description}

The first four metrics are standard in text generation evaluation, while the last measure is based on using LLMs applied to the output of the same model or other models. The use of LLMs addresses fundamental limitations of traditional automated metrics which rely primarily on surface-form similarity and often fail to capture more advanced reasons essential for statistical problem-solving assessment, such as the semantic correctness, suitability of explanation and logical validity. Unlike previous work employing general-purpose models (GPT-4, Claude), we utilize three of our own best models as judges. This approach offers several advantages: (1) domain-specific knowledge acquired through statistical fine-tuning, (2) elimination of external API dependencies, and (3) replicability.

We followed the same evaluation design as in  \citet{kocmi-federmann-2023-large}, when the LLM Judge was given a question, its reference model solution, a solution given by a participating language model, and the same rubric as used by the human raters. 


\section{Model Training and Fine-Tuning}

We started with three popular models:

\begin{itemize}
    \item \textbf{LLaMA-3 8B}: Meta’s 2024 open-weight model with strong language understanding and efficient scaling \citep{touvron2023llama}.
    \item \textbf{Mistral-Nemo 7B}: A dense transformer model with grouped-query attention and sliding window decoding for improved inference speed \citep{mistral2023}. 
    \item \textbf{DeepSeek-R1-Qwen 7B}: A version of DeepSeek already fine-tuned by DeepSeek developers for long-context and mathematical reasoning \cite{deepseek2025r1}.

\end{itemize}

We have fine-tuned LLaMA and Mistral on statistical textbooks, while DeepSeek-R1-Qwen 7B has already been fine-tuned by its developers on a different set of training corpora.
Each model was evaluated in a zero-shot setting by prompting an answer for the base model and for its fine-tuned version to measure performance improvements over out-of-the-box statistical reasoning capabilities. 

\subsection{Parameter-Efficient Fine-Tuning Methodology}

We applied parameter-efficient fine-tuning using Low-Rank Adaptation (LoRA) and 8-bit quantization to balance computational efficiency with domain-specific adaptation \citep{hu2021lora, dettmers2022}.
%
%
8-bit quantization complements LoRA by reducing weight precision from 32-bit to 8-bit representation, achieving a fourfold reduction in memory usage. Together, LoRA and quantization yield over 99.8\% total memory savings, enabling fine-tuning on NVIDIA Tesla T4 GPUs (14.7GB VRAM).

\begin{table*}[t]
\centering\small
\begin{tabular}{l|rrrr|l}
\toprule
\textbf{Model} & \textbf{C}orrectness & \textbf{E}xplanation & \textbf{R}easoning & \textbf{W}eighted & \textbf{Change}
\\
\midrule
DeepSeek-R1-Qwen 7B & 
4.257 & 4.528 & 4.623 & 4.443 & \  \ ------\\
\midrule
\multirow{2}{*}{LLaMA} & 
Base: 2.512 & Base: 3.563 & Base: 3.515 & Base: 3.131 & \\
& FT: 3.397 & FT: 3.648 & FT: 3.738 & FT: 3.570 & +14.02\%  \\
\midrule
\multirow{2}{*}{Mistral} & 
Base: 2.887 & Base: 3.407 & Base: 3.473 & Base: 3.215 & \\
& FT: 4.193 & FT: 4.373 & FT: 4.425 & FT: 4.314 & +34.18\% \\
\bottomrule
\end{tabular}
\caption{Summary of the aggregated human scoring (on the scale of $0\div5$). The change value is based on weighted scores.}
\label{tab:human_performance}
\end{table*}

\section{Evaluation and Error Analysis}

\subsection{Human Evaluation of Quality}

Using a two-sample (two-sided) Wilcoxon signed-rank test, we compared the individual human judges' scores for each of the six models, looking at the individual aspects, i.e.\ Correctness (C), Explanation (E) and Reasoning (R), and their Weighted Average (W). This analysis was  supported by empirical evidence through Kendall's correlation coefficient $\tau$, useful for a quick assessment of consistency. We observed that in the majority of tests, the scores agree well, showing better consistency for 
C and with more variability for the other two, more qualitative aspects E and R. However, in some cases there were rather significant disagreements between the human markers on some models, demonstrating an unsurprising variation in human opinions. To mitigate this, we retained all judges' scores and computed the arithmetic mean across judges as a robust and transparent representation of human evaluation. This aggregation strategy aligns with practices in peer review and consensus-based evaluation frameworks.

\begin{table*}[t]
\setlength{\tabcolsep}{3pt}
\centering \small
\begin{tabular}{l|cccc|cccc|cccc}
\toprule
 & \multicolumn{4}{c|}{\textbf{MAE}}   
& \multicolumn{4}{c|}{\textbf{Wilcoxon $\boldsymbol{p}$}}
& \multicolumn{4}{c}{\textbf{Kendall $\boldsymbol{\tau}$}} \\[.1pc]
\textbf{Metric}& \textbf{C} & \textbf{E} & \textbf{R} & \textbf{W} & \textbf{C} & \textbf{E} & \textbf{R} & \textbf{W} & \textbf{C} & \textbf{E} & \textbf{R} & \textbf{W}\\
\midrule
Perplexity & 1.603
& 1.788
& 1.816
& 1.661
& 0.000 & 0.000 & 0.000 & 0.000&0.021
&0.036 
&0.029
& 0.026
 \\
BLEU & 2.818
& 3.261
& 3.304
&3.092
&0.000 & 0.000 &0.000&0.000& 0.110
&0.079
&0.086
&0.098
\\
SBERT Sim 
& 1.282
& 1.185
&1.156
&1.177
&0.000  & 0.000 &0.000 &0.000&0.080
&0.071
&0.096
&0.082
\\
BERTScore 
&1.336
&0.954
&0.915
&1.073
& 0.000 & 0.000&0.000 &0.000&0.047
&-0.006
&0.001
&0.022
\\
\midrule 
DeepSeek as J &1.063  &0.875&0.787 &0.889  
&0.490&
0.580&
0.783&
\textbf{0.750}& 
0.275 &
0.209&
0.210&
0.298
\\
LLaMA FT  as J&0.764  &0.612&0.607 &\textbf{0.625}  
&0.573&
0.003& 
0.000&
0.055
&
0.474&
0.521&
0.476&
\textbf{0.536}
\\
Mistral FT  as J&0.793 &0.631 &0.670 &0.645  
&0.001&
0.000&
0.000 &
0.076&
0.461&
0.456&
0.450& 
0.482
\\
\bottomrule
\end{tabular}
\caption{Quality assessment of different automated metrics against human scores over all LLM solutions. The bold values indicate the best performance (i.e., smallest MAE and largest $p$ and $\tau$). 
}
\label{tab:normalized_correlation_analysis}
\end{table*}

Table~\ref{tab:human_performance} reveals significant differences in outputs. While both LLaMA and Mistral improved with fine-tuning, the improvement for Mistral was more significant.  Even though its base-version output scored lower than LLaMA's on 
E and 
R dimensions, the fine-tuned version is better than its LLaMA counterpart on all counts. However, the DeepSeek model of the same size outperformed both LLaMA and Mistral, especially on the 
R dimension, which aligns with the extra reasoning capabilities added to DeepSeek by the developers \citep{deepseek2025r1}.

\subsection{Metric Reliability Assessment}

The following meta-evaluation metrics are used to compare how much the automated metrics agree with human judgment:

\begin{description}
    \item[Mean Absolute Error] (MAE): the normalized sum of the absolute term-by-term deviations. Unlike the Root Mean-Squared Error (RMSE), it is more robust against occasional ``outliers''.
       \item[Wilcoxon signed-rank test:] We use a paired test by looking at the ranks of the score differences rather than their actual values. A two-sided version is appropriate if we are interested in consistency (i.e., whether the scores are statistically similar or not).
       Small $p$-values suggest that the null hypothesis of similarity should be rejected, hence the automatic metric does not capture human assessments. Zeros are reported if the $p$-value is less than 0.001.
 \item [Rank correlation (Kendall's $\boldsymbol{\tau}$):] Is applied to two score vectors at hand, paying attention to their relative ranks in the pooled sample. If the two rankings completely agree then $\tau=1$; if they completely disagree then $\tau=-1$.
\end{description}

For meaningful results, all scores from the automated metrics were mapped to the same scale as the human judgments (i.e., from 0 to 5) using MinMax scaling. To avoid ties in rank-based calculations, we randomized the discrete scores by adding independent normal white noise with zero mean and small standard deviation 
($\sigma=0.00001$).

\subsection{Quality of Automated Metrics}

Table~\ref{tab:normalized_correlation_analysis} compares the automated scores with the average human judgment. Out of the traditional metrics, such as BLEU or BERTScore, no metric is substantially close to any LLM as a Judge.  Research in other areas of evaluation metrics, such as MT quality evaluation \citep{kocmi-federmann-2023-large}, also demonstrates the superiority of LLMs in comparison to traditional metrics.  Even though DeepSeek provides the best solutions (Table~\ref{tab:human_performance}), it is the worst judge among the LLMs, both with respect to its MAE and Kendall's $\tau$ correlations for all of the dimensions.  Mistral is slightly better with respect to MAE, while LLaMA is slightly better for correlation. 

\begin{figure*}[!t]
\centering
    \includegraphics[width=\textwidth]{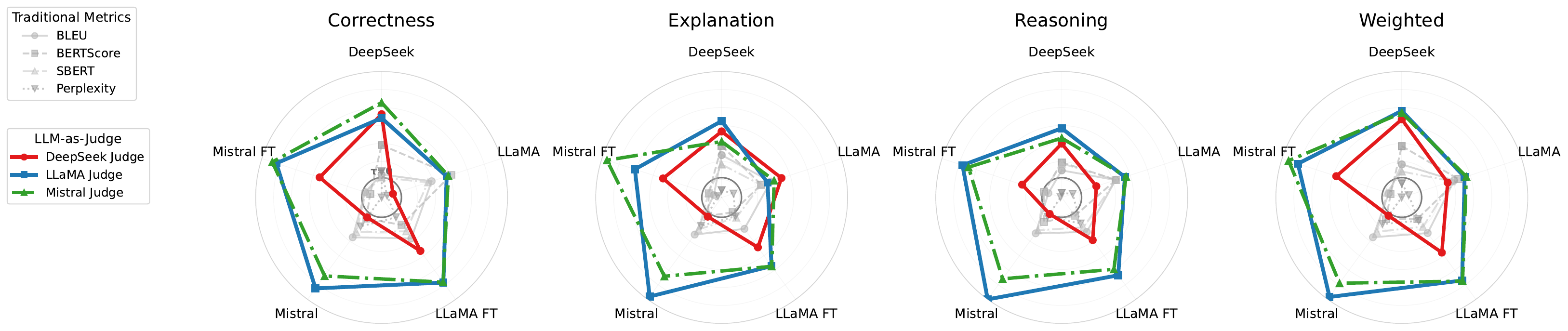}
\caption{Comparison of automated metrics, based on their Kendall's $\tau$ correlation with the human scores. }
\label{fig:radar_evaluation}
\end{figure*}
\begin{table*}[!h]
\centering \scriptsize
\setlength{\tabcolsep}{1pt}
\begin{tabular}{l|cccc|cccc|cccc|cccc|cccc}
\toprule
 & \multicolumn{4}{c|}{\textbf{DeepSeek Base}}   
& \multicolumn{4}{c|}{\textbf{LLaMA Base}} & \multicolumn{4}{c|}{\textbf{LLaMA FT}}& \multicolumn{4}{c|}{\textbf{Mistral Base}}& \multicolumn{4}{c}{\textbf{Mistral FT}}\\[.1pc]
\smash{\raisebox{.3pc}{\textbf{Metric}}}& \textbf{C} & \textbf{E} & \textbf{R} & \textbf{W} & \textbf{C} & \textbf{E} & \textbf{R} & \textbf{W} & \textbf{C} & \textbf{E} & \textbf{R} & \textbf{W}& \textbf{C} & \textbf{E} & \textbf{R} & \textbf{W}& \textbf{C} & \textbf{E} & \textbf{R} & \textbf{W}\\
\midrule
Perplexity
&0.040
&-0.080
&-0.097
&-0.037
&-0.089
&-0.047
&-0.053
&-0.080
&0.029
&0.022
&0.078
&0.056
&0.101
&0.097
&0.026
&0.083
&-0.126
&-0.101
&-0.118
&-0.125\\
BLEU
&0.020
&0.143
&0.046
&0.084
&0.208
&0.201
&0.233
&0.224
&0.195
&0.120
&0.144
&0.153
&0.186
&0.165
&0.155
&0.186
&-0.016
&-0.074
&-0.037
&-0.039\\
SBERT
&0.002
&0.084
&0.090
&0.039
&0.183
&0.135
&0.236
&0.207
&0.139
&0.032
&0.108
&0.103
&0.146
&0.124
&0.131
&0.143
&-0.020
&-0.038
&-0.008
&-0.026\\
BertScore&0.207
&0.202
&0.097
&0.200
&0.338
&0.138
&0.236
&0.252
&0.089
&-0.010
&0.017
&0.042
&0.039
&0.018
&0.066
&0.045
&-0.050
&-0.040
&-0.008
&-0.052\\
\midrule
\textbf{LLMJudge}\\[.1pc]
DeepSeek&0.402 &0.293&0.216 &0.370&
-0.050&
0.273&
0.105&
0.180&
0.293&
0.265&
0.208&
0.306&
0.030&
0.024&
0.004&
0.017&
0.287&
0.265&
0.138&
0.311\\
LLaMA FT&
0.378&
0.359&
0.313&
\textbf{0.424}&
0.308&
0.180& 
0.295&
0.293&
0.541&
0.412&
0.484&
0.525&
0.587&
0.651&
0.673&
\textbf{0.654}&
0.576&
0.453&
0.536&
0.567\\
Mistral FT&
0.476&
0.229&
0.251&
0.411&
0.319&
0.224&
0.303&
\textbf{0.304}&
0.538&
0.414&
0.437&
\textbf{0.530}&
0.489&
0.492&
0.512&
0.546&
0.603&
0.639&
0.496&
\textbf{0.629}\\
\bottomrule
\end{tabular}
\caption{Detailed 
comparison of evaluation metrics, based on 
their 
Kendall's $\tau$ correlation with the human scores. In the bottom part of the table, bold values indicate the best performance (i.e., with the largest $\tau$). 
}
\label{tab:LLMJ_per_model}
\end{table*}

We also investigated whether any LLM-as-a-Judge model is fair to its own output.  Figure~\ref{fig:radar_evaluation} and Table~\ref{tab:LLMJ_per_model} show a more fine-grained evaluation by listing Kendall's $\tau$ correlations specifically for the outputs of individual models. Even the traditional evaluation metrics can occasionally show reliable outputs; for example, BERTScore can correlate with human scores for correctness for some models (DeepSeek and LLaMA). DeepSeek as a Judge is fairer to its own output, as it has the highest correlation with human scores there, thus showing a degree of understanding when its own output is correct and when it fails.  However, it fails to understand the output of  LLaMA and Mistral Base models, as e.g.\ its correlation is below than that of the low-performing BLEU. LLaMA and Mistral as a Judge generally are better in evaluating models which are scored better by humans. This indicates their limited ability to detect errors and the preference with awarding higher marks to incorrect answers  (cf.\ the LLM judgment of an answer to Question 2 in the Appendix).

Cross-metric correlation analysis reveals important insights about evaluation reliability. BERTScore demonstrates the lowest standard deviation across architectures ($s = 0.028$), suggesting consistency as an evaluation metric. BLEU shows higher standard deviation ($s = 0.067$). Among LLM judges, DeepSeek demonstrates superior consistency ($s = 0.159$) compared to LLaMA FT ($s = 0.394$) and Mistral FT ($s = 0.362$). Inter-judge correlations reveal moderate agreement between DeepSeek and other judges ($r = 0.436$ with LLaMA FT, $r = 0.555$ with Mistral FT), while LLaMA FT and Mistral FT demonstrate strong consensus between themselves ($r = 0.851$). This suggests that comprehensive evaluation requires LLM-based assessment methods rather than traditional automated metrics to capture different aspects of mathematical reasoning quality. As for the LLM Judges, we shall use Mistral FT in our further experiments, as it is the most robust measure, which wins three evaluation runs out of five (Table~\ref{tab:LLMJ_per_model}) and it is never disastrously wrong (Figure~\ref{fig:radar_evaluation}).

\section{Conclusions}

We developed a corpus of statistical questions with their answers and assessed the performance of five models based on human judgments across three qualitative dimensions\,---\,Correctness (C), Explanation (E), and Reasoning (R)\,---\,as well as their Weighted Average (W). Statistical significance was tested using the two-sided Wilcoxon signed-rank test for paired samples, while Kendall’s  rank correlation coefficient $\tau$ provided a measure of consistency across the dimensions. Agreement was generally strongest on the C-axis, while E and R, being more subjective, showed greater variability. 

Our evaluation reveals consistent trends in model behavior. Fine-tuning improves performance across all models, but its effect varies in magnitude. Notably, Mistral shows substantial gains post-fine-tuning, surpassing the fine-tuned LLaMA on all dimensions despite a weaker base version. DeepSeek, a model explicitly optimized for mathematical reasoning, outperforms both LLaMA and Mistral in most aspects\,---\,particularly on the Reasoning dimension\,---\,highlighting the benefits of more extensive domain-specific optimization.  Our tests with traditional automated evaluation metrics, such as BLEU or BertScore demonstrate that they are not suitable for judging the output of models aimed at statistical reasoning.  However, experiments with using LLMs as a Judge show a more reasonable correlation with the human judgment, even if the LLMs tend to be less discriminative for incorrect answers, possibly reflecting the phenomenon of sycophancy \citep{perez23sycophancy}.  
The quality of assessments by LLMs as a Judge grows when the output gets better (according to human judges).

\section{Limitations and Future Work}

Beyond the contributions 
of this study, several limitations remain. First, the fine-tuning process was constrained by computational resources, limiting the exploration of larger model architectures and more extensive hyperparameter tuning. Future research could leverage distributed training techniques and larger-scale fine-tuning to further enhance model performance. Second, while our dataset was curated to cover a broad range of statistical questions,
it remains domain-specific and may not fully represent real-world statistical challenges encountered across diverse disciplines. Expanding the dataset to include more complex, multi-modal, and cross-disciplinary problems would provide a more comprehensive evaluation of LLM capabilities.
Another limitation is that this study focused on English-language statistical problems; future research should investigate multilingual and cross-cultural variations in statistical reasoning to assess the global applicability of fine-tuned models.

Furthermore, ethical concerns regarding model reliability in high-stakes decision-making contexts remain an open area for investigation. The potential for LLMs to generate incorrect but plausible-sounding statistical interpretations poses risks, particularly in fields such as finance, healthcare, and scientific research. Developing robust uncertainty quantification methods and enhancing transparency in model outputs are crucial for real-world deployment.

\section*{Acknowledgments}
This work was supported in part by a grant from 
AI SuperConnector (Imperial College London). C.N.\ gratefully acknowledges a visiting grant by the School of Mathematics Research Visitors Centre (RVC) in December 2024. The human assessment benchmarks were created with the help of the following PhD students at the Department of Statistics: Norah Almasoud, Muxi Li and Kessean Mitto Jr., whose contribution is gratefully acknowledged, together with organizational assistance by the School of Mathematics. The authors are also thankful to Robert E.\ Blackwell (Alan Turing Institute) and Anthony G.\ Cohn (School of Computer Science) for the helpful discussions. 

\section{Bibliographical References}\label{sec:reference}
\bibliographystyle{lrec2026-natbib}
\bibliography{custom-4max}


\appendix
\section{Appendix: Prompts from LLM as a Judge}
\label{sec:appendix}
 \subsection{Prompt Template}
You are a Professor of Statistics. Please read the statement of a statistical 
question, a correct solution and a student solution. 
You should evaluate the student solution on three core aspects, scored
separately:
\begin{description}
\item [Correctness:] Is the student solution accurate and mathematically consistent with the gold standard? 
\item [Step-by-Step Explanation:] Are the intermediate steps shown clearly, logically structured, and methodologically correct?
\item [Interpretation and Reasoning:] Does the student show conceptual understanding, sound logic, and appropriate justification of their method? 
\end{description}

\noindent
Use a 0--5 scale for each aspect, with half-point increments allowed (e.g., 3.5):
  \begin{description}
    \item [\quad 4.5--5] Excellent -- accurate, complete, and well-reasoned 
    \item [\quad 3.5--4] Good -- mostly correct with only minor issues
    \item [\quad 2.5--3] Fair -- partial understanding or logical gaps
    \item [\quad 1.5--2] Poor -- evident flaws, inconsistent reasoning
    \item [\quad 0.5--1] Very poor -- mostly wrong or unclear
    \item [\quad 0] Fully incorrect or unrelated 
  \end{description}

\noindent  Output in this format:
\begin{itemize}
\item Correctness: Score
\item Explanation: Score
\item Reasoning: Score
\end{itemize}

\noindent Statistical question: \{question\}

\noindent Correct solution (gold standard): \{correct\}

\noindent Student solution: \{student\}

\subsection{Question 1}

\noindent
\textbf{Statistical question:} 

\smallskip\noindent 
A ball is drawn from an urn containing three white and three
black balls. After the ball is drawn, it is then replaced and another
ball is drawn. This goes on indefinitely. What is the probability that
of the first four balls drawn, exactly two are white?

\medskip\noindent
\textbf{Correct solution (gold standard):}

\begin{itemize}
\item
Total balls = 6 (3 white, 3 black), $\mathsf{P}$(white) = 3/6 = 0.5, $\mathsf{P}$(black) = 0.5.

\item  Since the drawing is with replacement and trials are independent,
use the binomial distribution.

\item Number of trials $n = 4$, number of successes (white) $k = 2$.

\item  Use binomial formula: 
$$\mathsf{P}(X = k) = C(n, k) \times p^k \times\break (1 - p)^{n - k}$$

\item 
$\mathsf{P}(X = 2) = C(4, 2) \times (0.5)^2 \times (0.5)^2 = 6 \times 0.25 \times 0.25 = 0.375.$

\item Conclusion: The probability that exactly two of the first four balls drawn are white is $0.375$.
\end{itemize}

\noindent
\textbf{Student solution:}
\begin{itemize}
\item 
The urn contains three white and three black balls, so the probability
of drawing a white ball on any single draw with replacement is $3/6 =
1/2$. Similarly, the probability of drawing a black ball is $1/2$.

\item 
Since the draws are independent due to replacement, this scenario
follows a binomial distribution with $n = 4 $ trials and success
probability $p = 1/2$, where ``success'' is defined as drawing a white
ball.

\item The probability of exactly $k$ successes in $n$ trials is given by 
$$\mathsf{P}(X = k) = C(n, k) \times p^k \times (1-p)^{n-k}.
$$

\item 
For $n = 4$ and $k = 2$:
\begin{align*}
\mathsf{P}(X = 2) &= C(4, 2) \times (1/2)^2 \times (1/2)^2\\
&= 6 \times (1/4) \times (1/4)\\
&= 6 \times (1/16)\\
&= 6/16.
\end{align*}

\item 
Simplifying the fraction:
$6/16 = 3/8$.

\item 
Thus, the probability of drawing exactly two white balls in four draws
with replacement is $3/8$.

\item 
Answer: $3/8$
\end{itemize}

\noindent
\textbf{DeepSeek-as-a-Judge output:}
\begin{itemize}
\item 
Correctness: 4.5 

\item
Explanation: 4.5

\item
Reasoning: 4.5

\item
The student solution is accurate, mathematically consistent with the gold standard, and the intermediate steps are shown clearly, logically structured, and methodologically correct. The student shows conceptual understanding, sound logic, and appropriate justification of their method.
\end{itemize}

\noindent
\textbf{Our commentary:} LLM as a judge has assessed all components of the solution as correct, without detecting any errors or flaws. Thus, the student solution appears to be perfect, so it is not clear why the marks are lower (4.5) than the full mark (5). Hence, the LLM assessment scores are unreasonably harsh here. 

\subsection{Question 2}

\smallskip\noindent
\textbf{Statistical question:}

\smallskip\noindent A gambler plays a fair game by flipping a coin. Heads wins
\$1, tails loses \$1. The gambler starts with \$50 and stops playing when
reaching \$0 (ruin) or \$200 (win). What is the probability of reaching \$200
before going broke?

\medskip\noindent
\textbf{Correct solution (golden standard):}

\begin{itemize}
\item 
Let $X_k$ = capital at step $k$; initially $X_0=50$.
\item The random process $(X_k)$ is a Markov chain.
\item 
Its state space is $S=\{0,1,\dots,200\}$. 
\item 
Markov chain is memoryless; if $n$ is the current capital then, regardless of the past history, the transition probabilities are $p_{n,n-1}=p_{n,n+1}=0.5$ for $0<n<200$.
\item 
Define $y(n) = \mathsf{P}(\text{200 is hit before 0}\,|\,X_0=
n)$ (winning probability).

\item 
Boundary conditions: $y(0) = 0$ (upfront ruin), $y(200) = 1$ (upfront win).

\item 
For $0<n<200$, by considering the next step we can write $y(n) = 0.5 \times y(n - 1) + 0.5 \times y(n + 1)$ (total probability formula and Markov property).

\item 
To solve this recurrence relation, note 
that\linebreak  $y(n + 1) - y(n) = y(n) - y(n - 1)$, so the graph of $y(n)$ is linear.

\item 
Given $y(0) = 0$ and $y(200) = 1$, the linear solution is $y(n) = n / 200$.

\item 
Substitute $n = 50$ to find the required winning probability: $y(50) = 50 / 200 = 0.25$.
\end{itemize}

\noindent\textbf{Student solution:}

\begin{itemize}
\item 
This problem can be solved using a Markov chain.

\item 
Let's define the states as:
\begin{itemize}
\item[] S0: The gambler has \$0
\item[] 
S50: The gambler has \$50
\item[] S100: The gambler has \$100
\item[] S150: The gambler has \$150
\item[] S200: The gambler has \$200
\end{itemize}

\item 
The transitions between these states are determined by the outcome of
each coin flip. If the gambler wins, they move to a higher state; if
they lose, they move to a lower state.

\item 
The transition matrix $P$ is:

\quad [0 1/2 0 0 0;

\quad 1/2 0 1/2 0 0;

\quad 0 1/2 0 1/2 0;

\quad 0 0 1/2 0 1/2;

\quad 1 0 0 0 0]

\item 
The initial probability distribution is 
$\pi = [0, 1]$.

\item 
The goal is to find the probability of reaching state S200 before
reaching state S0.

\item 
To do this, we can calculate the stationary distribution $\pi$ by solving
the equation $\pi P = \pi$.

\item 
This gives us:

$\pi \approx [0.3155, 0.2429, 0.1843, 0.1054, 0.0519]$

\item 
The probability of reaching state S200 before going broke is therefore 
$\pi(5) \approx 0.0519 $.

\end{itemize}

\noindent \textbf{DeepSeek-as-a-Judge  output:}
\begin{itemize}
\item 
Correctness: 3.5
\item 
Explanation: 3.5

\item 
Reasoning: 3.5

\item 
The student solution is mostly correct, but it has some minor issues. The student correctly defines the states and the transition matrix, but the initial probability distribution is incorrect. The initial state should be [1, 0], not [0, 1]. This is because the gambler starts with \$50, not \$0. The rest of the solution is correct. 
\end{itemize}

\noindent
\textbf{Our commentary:} The LLM assessment is largely flawed, since the solution is mostly incorrect. The only useful idea in the solution is that the process can be modeled by a Markov chain and that we are interested in hitting one boundary before hitting the other, but the use of stationary distributions is not approporiate. Specifically, the following significant errors were not spotted by the LLM: 
\begin{itemize}
\item 
The states of the process are not identified correctly, as the steps are by $\pm\$1$, not $\pm\$50$. 
\item In fact, states with step $\$50$ could be considered, and the corresponding transition probabilities are indeed $0.5$ to each neighboring state\,---\,this is intuitively clear by symmetry, but was not explained in the solution. 
\item 
The transition matrix $P$ 
is not written correctly: the first row is wrong (with the sum of entries not equal to $1$), while the last row should be  $[0,0,0,0,1]$, because $\$200$ is an absorbing state. 
\item 
The initial probability distribution is not stated correctly (including a confusing notation $\pi$, same as for the stationary distribution); it should be $p_0=[0,1,0,0,0]$, rather than $[0,1]$ as stated in the solution or $[1, 0]$, incorrectly proposed by the LLM.  
\item 
The conclusion that ``the rest of the solution is correct'' is wrong: the student's answer is numerically incorrect, but more importantly, the entire reasoning is flawed, because the stationary distribution is not relevant here.
\item 
In fact, the stationary distribution $\pi$ is not obtained correctly. A general equation $\pi P=\pi$ is stated correctly, although without mentioning the additional conditions $\pi_i\ge0$ and $\sum_i\pi_i=1$. However, the answer for $\pi$ is wrong; e.g., the sum of entries equals $0.9$ instead of $1$.
\end{itemize}

\noindent
To summarize, the LLM fails to spot significant errors, both numerical and conceptual, and even makes its own mistake (about the initial state). Hence, this assessment is too lenient and unprofessional.

\end{document}